\documentclass[conference]{IEEEtran}
\IEEEoverridecommandlockouts
\usepackage{cite}
\usepackage{amsmath,amssymb,amsfonts}
\usepackage{algorithmic}
\usepackage{graphicx}
\usepackage{textcomp}
\usepackage{xcolor}
\usepackage[table]{xcolor}
\definecolor{taqgray}{gray}{0.90}
\usepackage{booktabs}
\usepackage{multirow}
\def\BibTeX{{\rm B\kern-.05em{\sc i\kern-.025em b}\kern-.08em
    T\kern-.1667em\lower.7ex\hbox{E}\kern-.125emX}}
\begin{document}

\title{PersonalQ: Select, Quantize, and Serve Personalized Diffusion Models for Efficient Inference
}

\author{
\IEEEauthorblockN{
\textbf{Qirui Wang}$^{1*}$, 
\textbf{Qi Guo}$^{1*}$, 
\textbf{Yiding Sun}$^{1}$, 
\textbf{Junkai Yang}$^{1}$, 
\textbf{Dongxu Zhang}$^{1}$, 
\textbf{Shanmin Pang}$^{1\dagger}$, 
\textbf{Qing Guo}$^{2\dagger}$
}
\IEEEauthorblockA{
$^{1}$School of Software Engineering, Xi'an Jiaotong University, China \\
$^{2}$Nankai University, China
}
}

\maketitle
{
  \renewcommand{\thefootnote}{\fnsymbol{footnote}}
  \footnotetext[1]{Equal contribution.}
  \footnotetext[2]{Corresponding authors.}
}

\begin{abstract}
Personalized text-to-image generation lets users fine-tune diffusion models into repositories of concept-specific checkpoints, but serving these repositories efficiently is difficult for two reasons: natural-language requests are often ambiguous and can be misrouted to visually similar checkpoints, and standard post-training quantization can distort the fragile representations that encode personalized concepts. We present \textbf{PersonalQ}, a unified framework that connects checkpoint selection and quantization through a shared signal---the checkpoint’s \emph{trigger token}. \emph{Check-in} performs intent-aligned selection by combining intent-aware hybrid retrieval with LLM-based reranking over checkpoint context and asks a brief clarification question only when multiple intents remain plausible; it then rewrites the prompt by inserting the selected checkpoint’s canonical trigger. Complementing this, \emph{Trigger-Aware Quantization} (TAQ) applies trigger-aware mixed precision in cross-attention, preserving trigger-conditioned key/value rows (and their attention weights) while aggressively quantizing the remaining pathways for memory-efficient inference. Experiments show that PersonalQ improves intent alignment over retrieval and reranking baselines, while TAQ consistently offers a stronger compression--quality trade-off than prior diffusion PTQ methods, enabling scalable serving of personalized checkpoints without sacrificing fidelity.
\end{abstract}


\section{Introduction}
Diffusion-based~\cite{sohl2015deep} text-to-image models~\cite{podell2023sdxl} are commonly personalized via DreamBooth~\cite{ruiz2023dreambooth} and LoRA~\cite{hu2022lora}. They produce \emph{personalized checkpoints}: weights that respond to trigger tokens and reproduce a specific object, character, or style (e.g., a user’s teddy bear bound to \texttt{<bear-v4>}).

\begin{figure}[t]
\centering
\includegraphics[width=1\columnwidth]{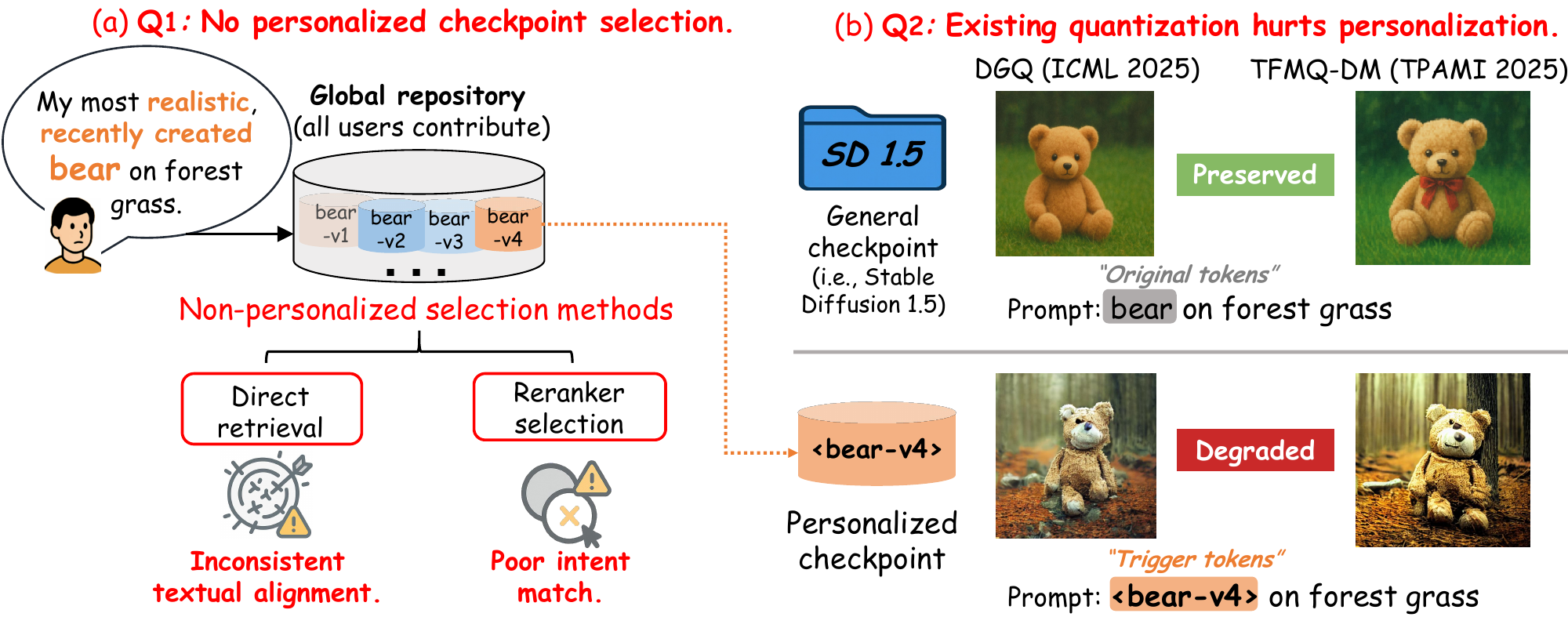}
\caption{
\textbf{Problem and overview.}
(a) Natural-language prompts must be routed to the correct personalized checkpoint; existing methods often misroute queries.
(b) Standard post-training quantization degrades personalized concepts.}
\label{fig:problem}
\end{figure}

As users keep fine-tuning, they accumulate repositories with many checkpoints, including multiple versions and temporally separated variants of the same concept. At inference, users refer to these latent choices in natural language (e.g., ``my most realistic, recently created bear on forest grass''). Selection is often manual or based on non-personalized retrieval~\cite{luo2024stylus}, which can misroute prompts when concepts are visually similar or descriptions overlap (Figure~\ref{fig:problem}a). \textbf{Q1:} How can we reliably map a user prompt to the \emph{intended} checkpoint in a large repository?

Even with the correct checkpoint, serving many personalized models is memory-intensive: each request may require loading its checkpoint weights onto the GPU, quickly exhausting capacity under concurrency (e.g., $\sim$4\,GB GPU memory per model). Quantization helps, but standard PTQ methods~\cite{li2023qdiffusion,huang2024tfmqdm,ryu2025dgq} can disrupt the delicate pathways encoding personalized concepts, harming text--image alignment and identity fidelity (Figure~\ref{fig:problem}b). \textbf{Q2:} How can we reduce GPU memory for serving personalized diffusion models while preserving personalization quality?

We view these questions as two facets of a single problem: \emph{intent-aligned, memory-efficient serving of personalized diffusion repositories}. We introduce \textbf{PersonalQ}, a unified framework that links checkpoint selection and quantization through a shared signal: \emph{trigger tokens}. In PersonalQ, trigger tokens simultaneously (i) define the personalized concepts needed by our selection module, \emph{Check-in}, and (ii) identify the most fragile pathways under quantization, which our \emph{Trigger-Aware Quantization} (TAQ) module explicitly protects. In this way, the same signals that specify \emph{which checkpoint} to serve guide \emph{what information} must be preserved.

We further introduce \textsc{Repo-Prompts}, a benchmark that mixes natural content descriptions with repository context cues to evaluate checkpoint selection in
large personalized repositories. We construct a repository of 1{,}000 personalized text-to-image
checkpoints spanning 20 concept categories with 50 temporal versions each, and curate
\textsc{Repo-Prompts} with 500 user-like queries, including single-match requests, ambiguous prompts
that require clarification, and no-match cases. On this setup, \emph{Check-in} achieves the highest
LLM-judge \emph{Intent Score} of $4.42 \pm 0.51$ and is consistently preferred by human raters,
winning $89.1\%$, $85.7\%$, and $82.1\%$ of pairwise comparisons against \emph{Random},
\emph{Reranker}, and \emph{Stylus}, respectively. On automatic benchmarks (MS-COCO and
PartiPrompts) across Stable Diffusion v1.5 and SDXL-Turbo, \emph{Trigger-Aware Quantization} (TAQ)
yields the strongest compression--quality trade-off: it stays close to full precision in the 8-bit
setting and remains robust under more aggressive activation quantization, while substantially
reducing inference memory (4--8$\times$) and computation in bit-operations (16--32$\times$).

In summary, this paper makes the following contributions:
\begin{itemize}
    \item We introduce \emph{Check-in}, a module that
    selects a personalized checkpoint from a large repository using a natural-language request. It
    jointly reasons over visual descriptors and metadata (e.g., style cues and temporal/version
    references), supports clarification for ambiguous intents, and outputs a trigger-tokenized prompt
    targeted to the selected checkpoint.

    \item We present \emph{Trigger-Aware Quantization} (TAQ), a post-training quantization method that preserves
    precision along trigger-token cross-attention pathways while aggressively quantizing remaining
    components, enabling smaller inference memory and 16--32$\times$ fewer bit-operations
    with strong fidelity under low-bit inference.

    \item We release a 1{,}000-checkpoint personalized
    repository and the \textsc{Repo-Prompts} benchmark for context-aware selection. Together, they
    provide a controlled testbed for large-scale personalized checkpoint retrieval, clarification,
    and efficient serving under quantization.
\end{itemize}

\begin{figure}[t]
  \centering
  \includegraphics[width=1\columnwidth]{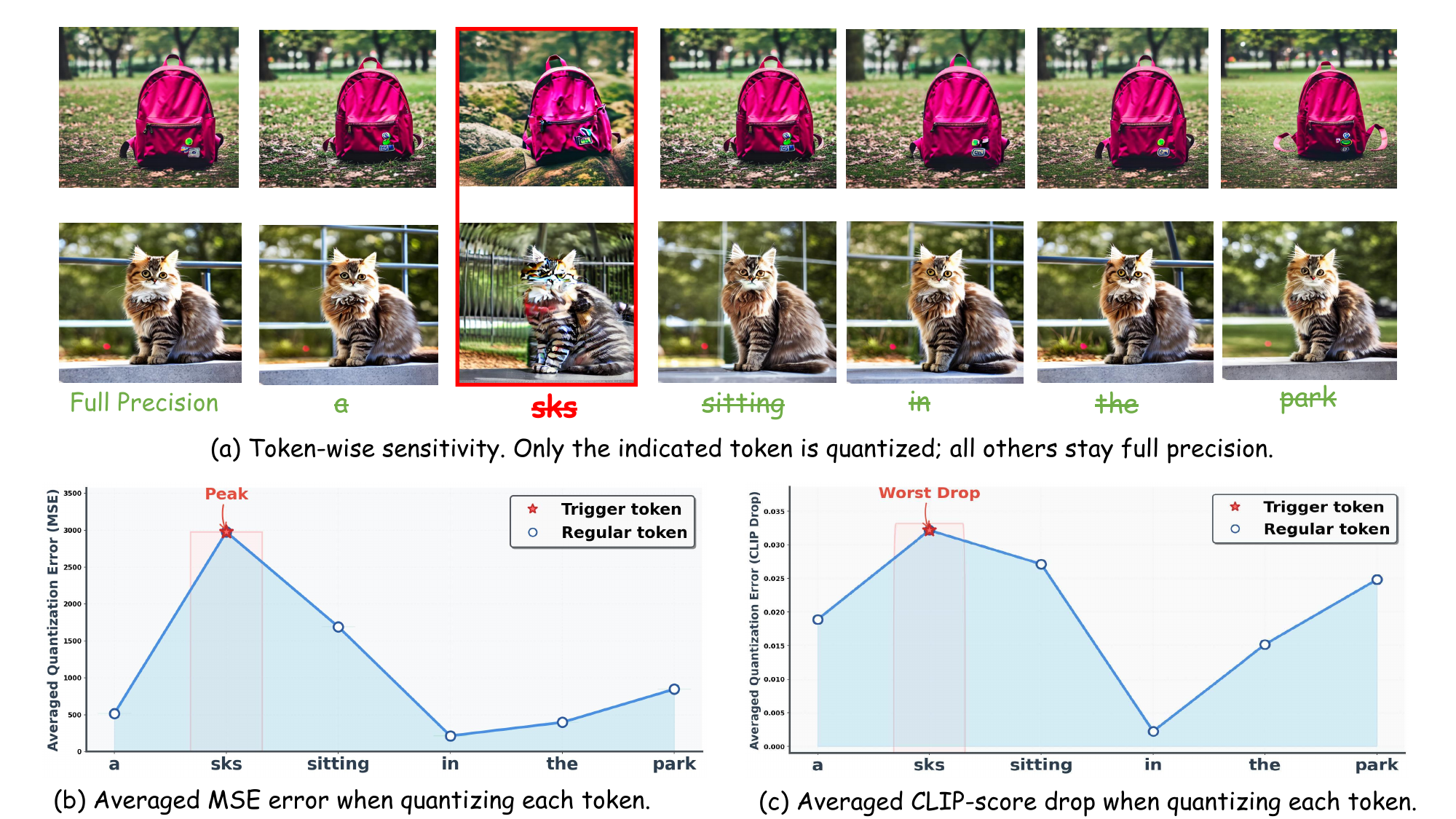}
  \caption{\textbf{Trigger token (e.g., \texttt{<sks>}) is vulnerable under quantization.} Following the token-specific sensitivity test in Sec.~\ref{sec:TAQ}, we quantize only the cross-attention Key/Value rows of the trigger token while keeping all other rows in full precision. Under 4-bit quantization, the trigger token exhibits much larger degradation than ordinary words, motivating our \emph{TAQ} method.}
  \label{fig:error}
\end{figure}

\section{Related work}
\subsection{Checkpoint Selection and Management}
Existing selection methods like Stylus \cite{luo2024stylus} match artistic styles for LoRA selection, while Mix-of-Show \cite{gu2023mix} merges adapters; broader efforts on efficient adaptation and multimodal visual representation learning \cite{sun2026align, SUN2026112800} likewise improve transferable. However, these methods assume users specify exact needs and ignore practical constraints like version history and memory limits. Standard retrieval methods (RAG \cite{ zhu2024acceleratinginferenceretrievalaugmentedgeneration}, rerankers \cite{ma2023large}) handle text well but struggle to reason over checkpoint context attributes and ambiguous requests like "use the latest cat version with artistic style." While reranking approaches \cite{niu2024judgerank} use language model reasoning, our \emph{Check-in} uniquely combines prompt understanding with personalized reasoning over checkpoint context to resolve ambiguous requests within resource constraints.
\subsection{Post-training Quantization of Diffusion Models}
Post-training quantization (PTQ) reduces model size by using low-bit representations (e.g., 4-8 bits). Diffusion models are challenging to quantize because activation ranges vary significantly across timesteps \cite{wang2024towards}. Existing PTQ methods use timestep-adaptive strategies \cite{li2023qdiffusion, huang2024tfmqdm}, preserve outliers \cite{ryu2025dgq}, or more broadly adapt computation across inputs \cite{zhang2026not}. Our TAQ method specifically targets personalization by maintaining high precision for trigger-token pathways while aggressively quantizing other components, preserving personalization quality under memory constraints.

\begin{figure*}[t]
  \centering
  \includegraphics[width=0.8\textwidth]{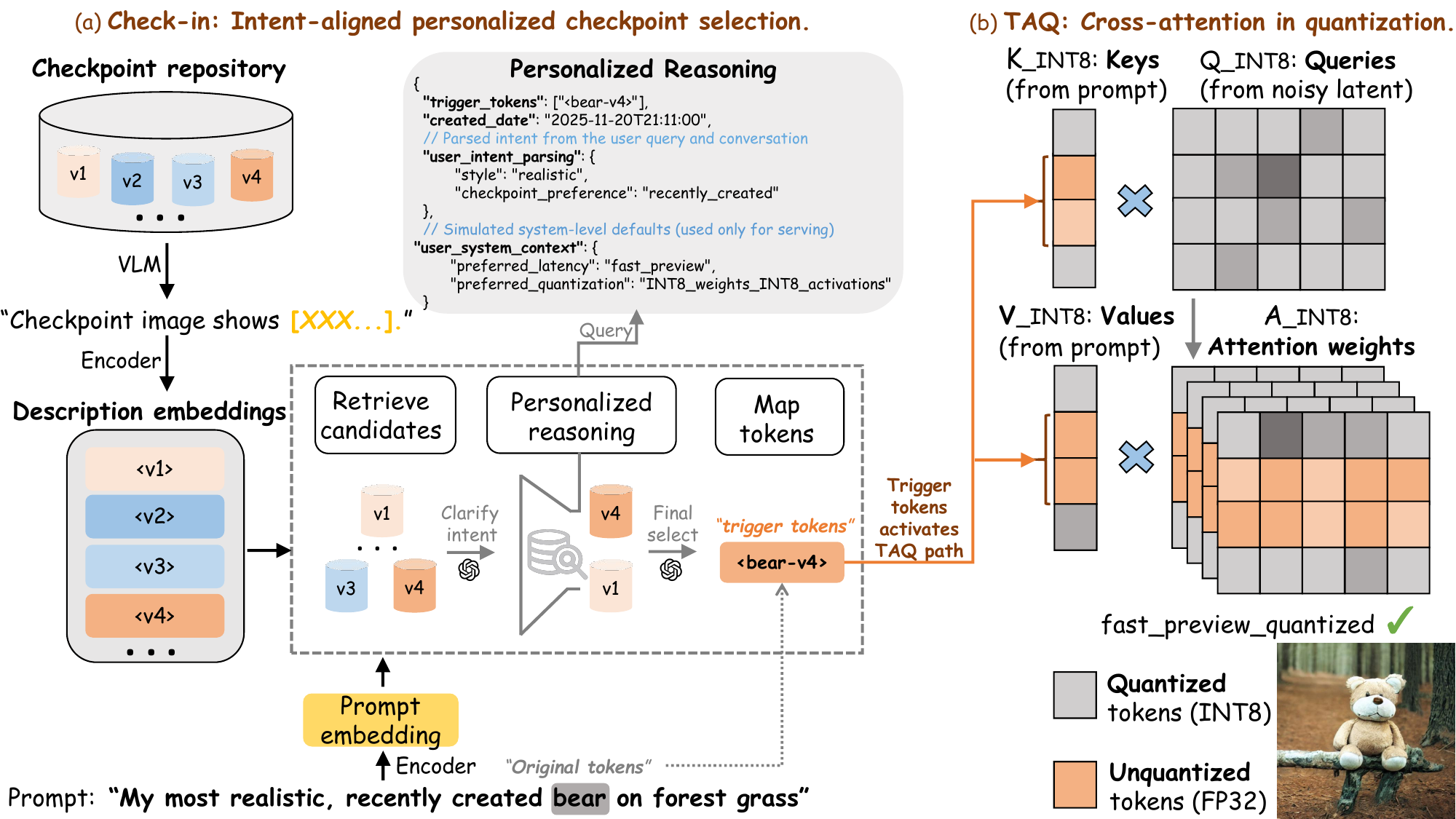}
  \caption{\textbf{PersonalQ.} \textbf{Input:} prompt $p$ and a personalized checkpoint repository. \textbf{Output:} image $\hat{\mathbf{y}}$ from low-bit inference.
  (a) \emph{Check-in} selects a checkpoint $c^\star$ using prompt semantics and metadata, then rewrites $p$ into $p'$ by inserting the checkpoint’s trigger (e.g., \texttt{bear}$\rightarrow$\texttt{\textless bear-v4\textgreater}).
  (b) \emph{TAQ} quantizes cross-attention with trigger-aware mixed precision: it keeps trigger-conditioned K/V rows (and their attention pathways) in higher precision and quantizes the remaining activations for efficient inference with preserved personalization.}
  \label{fig:method}
\end{figure*}

\section{Methodology}
We propose a two-part system for accurate selection and efficient, fidelity-preserving inference over repositories of personalized checkpoints: \emph{(A) Check-in} reasons in a personalized checkpoint attribute space built from repository context and prompt semantics, maps generic nouns to trigger tokens, and resolves ambiguity (Sec.\ref{sec:checkin}); and \emph{(B) TAQ}—Trigger-Aware quantization that preserves identity-critical pathways while quantizing the rest (Sec.\ref{sec:TAQ}). Together, the components preserve user intent in checkpoint selection, maintain subject fidelity under quantization, and meet throughput and memory targets, enabling scalable personalized image generation. Figure~\ref{fig:method} illustrates our pipeline.

\subsection{\emph{Check-in}: Intent-aligned personalized checkpoint selection}
\label{sec:checkin}

We assume a repository of pre-trained personalized checkpoints
$\mathcal{C}=\{c_1,\dots,c_n\}$.
Users provide a text-only prompt $p$ (e.g., ``my most realistic, recently created bear on forest grass'').
The system internally selects a checkpoint $c^\star \in \mathcal{C}$ that best matches the user’s intent, then rewrites the prompt to include the corresponding trigger token for downstream quantized inference (see Sec.~\ref{sec:TAQ}).

For each checkpoint, the training pipeline renders a small set of visual previews.
A vision--language model summarizes these previews into a text description $D_i$.
We encode $D_i$ into an embedding $e_i = E(D_i)$ and store $\{e_i\}_{i=1}^n$ for candidate retrieval in Step~1.

Accordingly, each checkpoint $c_i$ is represented as a context tuple
$c_i=(T_i, S_i^{\text{subj}}, S_i^{\text{style}}, D_i, M_i)$:
\begin{itemize}
  \item $T_i$: trigger tokens (e.g., \texttt{<bear-v4>}).
  \item $S_i^{\text{subj}}$: subject types (e.g., \textit{bear}).
  \item $S_i^{\text{style}}$: style tags (e.g., \textit{realistic}).
  \item $D_i$: a natural-language description summarizing the visual previews.
  \item $M_i$: context attributes such as \texttt{created\_at} and optional version identifiers.
\end{itemize}

\textbf{Step 1: Intent-aware hybrid retrieval.}
We build a sparse ``checkpoint card''
$X_i = \text{concat}(S_i^{\text{subj}}, S_i^{\text{style}})$, which provides exact noun and style anchors.
Given a user prompt $p$, we compute dense and sparse similarities jointly:
\[
\begin{bmatrix}
\text{sim}_{\text{dense}}(p,c_i)\\
\text{sim}_{\text{sparse}}(p,c_i)
\end{bmatrix}
=
\begin{bmatrix}
\frac{E(p)\cdot e_i}{\|E(p)\|\,\|e_i\|}\\
\mathrm{BM25}(p, X_i)
\end{bmatrix}.
\]
We then fuse the dense and sparse rankings with Reciprocal Rank Fusion (RRF)~\cite{cormack2009reciprocal}:
\[
\text{score}(p,c_i)=\sum_{m\in\{\text{dense},\text{sparse}\}} \frac{1}{\kappa + r_m(c_i)},
\]
where $r_m(c_i)$ denotes the rank of $c_i$ under modality $m$ and we use $\kappa{=}60$.
We keep the top-$K$ checkpoints (we use $K{=}10$) as candidates $\mathcal{C}_K$.

From $p$ and user-preferred inference settings, we construct a compact \emph{intent record} that contains
(i) \emph{user intent} inferred from the current query and
(ii) \emph{system context} inferred from the user’s prior history.
In our experiments, we do not use real user histories; instead, we simulate a fixed system context by assigning default values to each user, and use it only to configure serving behavior after checkpoint selection. If multiple plausible intents would lead to different checkpoints, we ask a short clarification using the most discriminative attribute(s), e.g.,
\emph{``Do you want a cute stuffed teddy bear, or a realistic brown bear?''}
The answer updates the intent record and resolves ambiguity without exposing checkpoint identifiers.

\textbf{Step 2: Personalized reasoning for reranking and serving.}
Given the intent record, we rerank the candidate set $\mathcal{C}_K$ using only the \emph{user-intent fields}:
semantic/style match against each checkpoint description ($D_i$) and recency signals from $M_i$
(e.g., \texttt{created\_at} combined with query cues such as \texttt{recently\_created}).
The \emph{system-default fields} are not used as selection constraints; instead, they determine \emph{how} to serve the chosen checkpoint,
such as selecting a fast-preview quantization setting.
We implement this step with Gemini 2.5 Flash~\cite{team2023gemini}.

\textbf{Step 3: Map original tokens to trigger tokens.}
After selecting checkpoint $c^\star$, we rewrite the prompt to include its canonical trigger token.
For each noun $w$ in $p$ that matches the selected subject types $S_{c^\star}^{\text{subj}}$, we apply
\texttt{$p'=\mathrm{map}(p,w,t_{c^\star})$},
where $t_{c^\star}\in T_{c^\star}$ is the canonical token.
For example, mapping \texttt{bear} to \texttt{<bear-v4>} yields
\emph{``my most realistic, recently created \texttt{<bear-v4>} on forest grass''}.
We pass $p'$ to the downstream inference pipeline.

\subsection{Trigger-Aware Quantization: Addressing the Challenge}
\label{sec:TAQ}

\textbf{Why personalized diffusion models are different.}
A personalized checkpoint binds a learned \emph{trigger token} (e.g., \texttt{<sks>}) to a user-specific concept. Depending on the tokenizer, this trigger may be represented by a \emph{contiguous} span of sub-tokens with indices $\mathcal{I}_{\text{sks}}$ (span $\mathcal{S}$); all remaining text-token indices are $\mathcal{I}_{\text{other}}$.

In each cross-attention head, queries $\mathbf{Q}$ attend to text keys/values $\mathbf{K},\mathbf{V}$ (one row per text token), with attention
\[
\mathbf{A}=\operatorname{softmax}\!\left(\mathbf{Q}\mathbf{K}^{\top}/\sqrt{d}\right),
\]
where $d$ is the key dimension. Therefore, the trigger influences the image only through the $\mathcal{I}_{\text{sks}}$ rows of $\mathbf{K}$ and $\mathbf{V}$ (and the corresponding columns of $\mathbf{A}$); distorting these rows directly degrades how the personalized concept is rendered.

\textbf{Token-specific quantization sensitivity.}
To isolate which tokens are most vulnerable, we run a surgical experiment:
instead of quantizing the entire model, we quantize \emph{only} the
$\mathbf{K}$ and $\mathbf{V}$ rows corresponding to a chosen token (or
trigger span) $\mathcal{S}$ to $b$ bits (using a standard affine quantizer
$\mathcal{Q}_b$), while keeping all other parameters and activations in
full precision. For each token $i$, we run inference with only token $i$
quantized and obtain an output $\mathbf{y}^{(i,b)}$.

Given a dissimilarity metric $\mathcal{L}$ (smaller is better), we define
the per-token sensitivity as
\begin{equation}
\Delta_i(b) = \mathcal{L}\big(\mathbf{y}^{(i,b)}, \mathbf{y}^\star\big),
\qquad b \in \{8,4\},
\label{eq:delta}
\end{equation}
where $\mathbf{y}^\star$ is the full-precision output.
We use two complementary metrics: (1) image MSE to capture visible
degradations (blur, artifacts), and (2) CLIP score~\cite{radford2021learning}
to detect semantic drift. Finally, we aggregate $\Delta_i(b)$ over $i \in \mathcal{I}_{\text{sks}}$
and $i \in \mathcal{I}_{\text{other}}$. This reveals that trigger tokens are
disproportionately sensitive to quantization, and Fig.~\ref{fig:error}
illustrates the procedure and shows that the trigger token \texttt{<sks>}
is substantially more fragile than common words under 4-bit quantization.

\textbf{Trigger-Aware Quantization (TAQ).}
Motivated by this, TAQ adopts a selective strategy: preserve
personalized trigger pathways in high precision and aggressively
quantize the rest.

We construct binary masks that mark trigger positions. For keys and
values, $\mathbf{M}_{\text{KV}} \in \{0,1\}^{H \times T \times 1}$
is broadcast across batch, layers, and channels; for attention,
$\mathbf{M}_{\text{A}} \in \{0,1\}^{B \times N \times T}$ marks the
same token indices. Mask entries are $1$ at trigger indices
($i \in \mathcal{I}_{\text{sks}}$) and $0$ elsewhere, and for multi-token
triggers the entire span $\mathcal{S}$ is protected.

TAQ then applies low-bit quantization $\mathcal{Q}_a(\cdot)$ only to
non-trigger elements:
\begin{equation}
\label{eq:TAQ_kvq}
\begin{aligned}
\tilde{\mathbf{K}} &=
  \mathbf{M}_{\text{KV}}\odot\mathbf{K}
  + (1-\mathbf{M}_{\text{KV}})\odot \mathcal{Q}_a(\mathbf{K}),\\
\tilde{\mathbf{V}} &=
  \mathbf{M}_{\text{KV}}\odot\mathbf{V}
  + (1-\mathbf{M}_{\text{KV}})\odot \mathcal{Q}_a(\mathbf{V}),\\
\tilde{\mathbf{Q}} &=
  \mathcal{Q}_a(\mathbf{Q}),
\end{aligned}
\end{equation}
\begin{equation}
\label{eq:TAQ_attention}
\hat{\mathbf{A}} =
  \mathbf{M}_{\text{A}} \odot \mathbf{A}
  + (1 - \mathbf{M}_{\text{A}}) \odot \mathcal{Q}_a(\mathbf{A}).
\end{equation}
In words, TAQ keeps all trigger-related K/V rows and their attention
weights in full precision, while quantizing all remaining text and
image pathways. This preserves personalized concepts with minimal
quality loss, yet still achieves substantial memory savings for
deployment.

\begin{table}[t]
\caption{\textbf{Quantitative Comparison.} Results of different quantization methods for personalized diffusion checkpoints.}
\label{tab:main}
\centering
\resizebox{\linewidth}{!}{%
\setlength{\tabcolsep}{4pt}
\renewcommand{\arraystretch}{1.1}
\begin{tabular}{llcc|cc|cc}
\toprule[1pt]
\multirow{2}{*}{\textbf{\shortstack{Checkpoint\\Model}}} &
\multirow{2}{*}{\textbf{\shortstack{Quantization\\Method}}} &
\textbf{Bit-width} & \textit{BOPs} &
\multicolumn{2}{c}{\textbf{MS-COCO}} &
\multicolumn{2}{c}{\textbf{PartiPrompts}} \\
& & (W/A) & \textit{(T)} &
\textbf{FID$\downarrow$} & \textbf{CLIP$\uparrow$} &
\textbf{FID$\downarrow$} & \textbf{CLIP$\uparrow$} \\
\midrule

\multirow{9}{*}{\shortstack{Stable\\Diffusion-v1-5}}
& Full Precision   & 32/32 & 893  & 10.96 & 0.315 & 9.77  & 0.336 \\
\cmidrule(lr){2-8}

& Q-Diffusion     & 8/8   & 56   & 27.16 & 0.261 & 23.44 & 0.267 \\
& TFMQ-DM         & 8/8   & 56   & 24.34 & 0.279 & 21.66 & 0.299 \\
& DGQ             & 8/8   & 56   & 15.24 & 0.291 & 13.26 & 0.317 \\
& \cellcolor{taqgray}\textbf{\textit{TAQ}}
  & \cellcolor{taqgray}8/8
  & \cellcolor{taqgray}56
  & \cellcolor{taqgray}\textbf{11.03}
  & \cellcolor{taqgray}\textbf{0.297}
  & \cellcolor{taqgray}\textbf{10.49}
  & \cellcolor{taqgray}\textbf{0.327} \\
\cmidrule(lr){2-8}

& Q-Diffusion     & 8/4   & 28   & 223.11 & 0.082 & 218.93 & 0.084 \\
& TFMQ-DM         & 8/4   & 28   & 167.22 & 0.138 & 173.51 & 0.141 \\
& DGQ             & 8/4   & 28   & 53.39  & 0.248 & 55.13  & 0.244 \\
& \cellcolor{taqgray}\textbf{\textit{TAQ}}
  & \cellcolor{taqgray}8/4
  & \cellcolor{taqgray}28
  & \cellcolor{taqgray}\textbf{38.84}
  & \cellcolor{taqgray}\textbf{0.264}
  & \cellcolor{taqgray}\textbf{36.91}
  & \cellcolor{taqgray}\textbf{0.265} \\

\midrule

\multirow{9}{*}{\shortstack{Stable\\Diffusion-XL\\Turbo}}
& Full Precision   & 32/32 & 6930 & 15.25 & 0.305 & 13.62 & 0.312 \\
\cmidrule(lr){2-8}

& Q-Diffusion     & 8/8   & 433  & 108.81 & 0.091 & 102.54 & 0.094 \\

& TFMQ-DM         & 8/8   & 433  & 78.63 & 0.151 & 72.81 & 0.164 \\
& DGQ             & 8/8   & 433  & 33.12  & 0.252 & 30.28  & 0.258 \\
& \cellcolor{taqgray}\textbf{\textit{TAQ}}
  & \cellcolor{taqgray}8/8
  & \cellcolor{taqgray}433
  & \cellcolor{taqgray}\textbf{22.18}
  & \cellcolor{taqgray}\textbf{0.293}
  & \cellcolor{taqgray}\textbf{21.37}
  & \cellcolor{taqgray}\textbf{0.294} \\
\cmidrule(lr){2-8}

& Q-Diffusion     & 8/4   & 216  & 153.35 & 0.032 & 147.42 & 0.043 \\
& TFMQ-DM         & 8/4   & 216  & 111.59 & 0.145 & 107.91 & 0.138 \\
& DGQ             & 8/4   & 216  & 52.73  & 0.229 & 47.58  & 0.223 \\
& \cellcolor{taqgray}\textbf{\textit{TAQ}}
  & \cellcolor{taqgray}8/4
  & \cellcolor{taqgray}216
  & \cellcolor{taqgray}\textbf{34.15}
  & \cellcolor{taqgray}\textbf{0.254}
  & \cellcolor{taqgray}\textbf{30.46}
  & \cellcolor{taqgray}\textbf{0.227} \\

\bottomrule[1pt]
\end{tabular}%
}
\end{table}

\subsection{Experimental Setup}
\label{subsec:exp-setup}

\textbf{Personalized checkpoints.}
We build a repository of 1{,}000 personalized Stable Diffusion checkpoints over 20 concept categories and 50 temporal versions per category (v1--v50): \texttt{<dog>}, \texttt{<cat>}, \texttt{<person>}, \texttt{<bear>}, \texttt{<horse>}, \texttt{<car>}, \texttt{<toy>}, \texttt{<watch>}, \texttt{<bag>}, \texttt{<chair>}, \texttt{<house>}, \texttt{<building>}, \texttt{<bridge>}, \texttt{<flower>}, \texttt{<tree>}, \texttt{<mountain>}, \texttt{<painting>}, \texttt{<drawing>}, \texttt{<logo>}, and \texttt{<shoe>}. Each checkpoint is tuned from 3--5 concept images. For SD1.5, we use DreamBooth with AdamW, 512 resolution, learning rate $5\times10^{-6}$, 600 steps, seed 42, and prior preservation; for SDXL-Turbo, we use LoRA DreamBooth with AdamW, 1024 resolution, learning rate $3\times10^{-5}$, 400 steps, seed 42, and prior preservation.

\textbf{Repo-Prompts dataset.}
We introduce \textsc{Repo-Prompts}, a dataset of 500 natural-language queries targeting specific checkpoints (e.g., ``the April dog model'', ``latest version'', ``anime style from last week''). Each instance contains \texttt{instance\_id}, \texttt{query}, \texttt{candidate\_pool}, and \texttt{ground\_truth} with \texttt{checkpoint\_id}, \texttt{requires\_clarification}, and \texttt{no\_match}. The dataset contains 350 single-match, 100 ambiguous, and 50 no-match queries.

\textbf{Baselines.}
For quantization, we compare Q-Diffusion~\cite{li2023qdiffusion}, TFMQ-DM~\cite{huang2024tfmqdm}, and DGQ~\cite{ryu2025dgq}. These baselines do not perform intent-aligned checkpoint selection; all methods use the same checkpoints and prompts. For selection, we include \emph{Random}, \emph{Reranker} (Qwen3-Reranker-4B~\cite{zhang2025qwen3}), and \emph{Stylus}~\cite{luo2024stylus}.

\textbf{Weight quantization.}
We also evaluate weight PTQ using the same pipeline across methods, reporting Adaround~\cite{nagel2020up} and BRECQ~\cite{li2021brecq}. All methods use the same 64 MS-COCO captions for calibration and reconstruction; block reconstruction is applied to both transformer and residual blocks using intermediate outputs from the same set.

\begin{figure*}[!t]
\centering
\includegraphics[width=1\columnwidth]{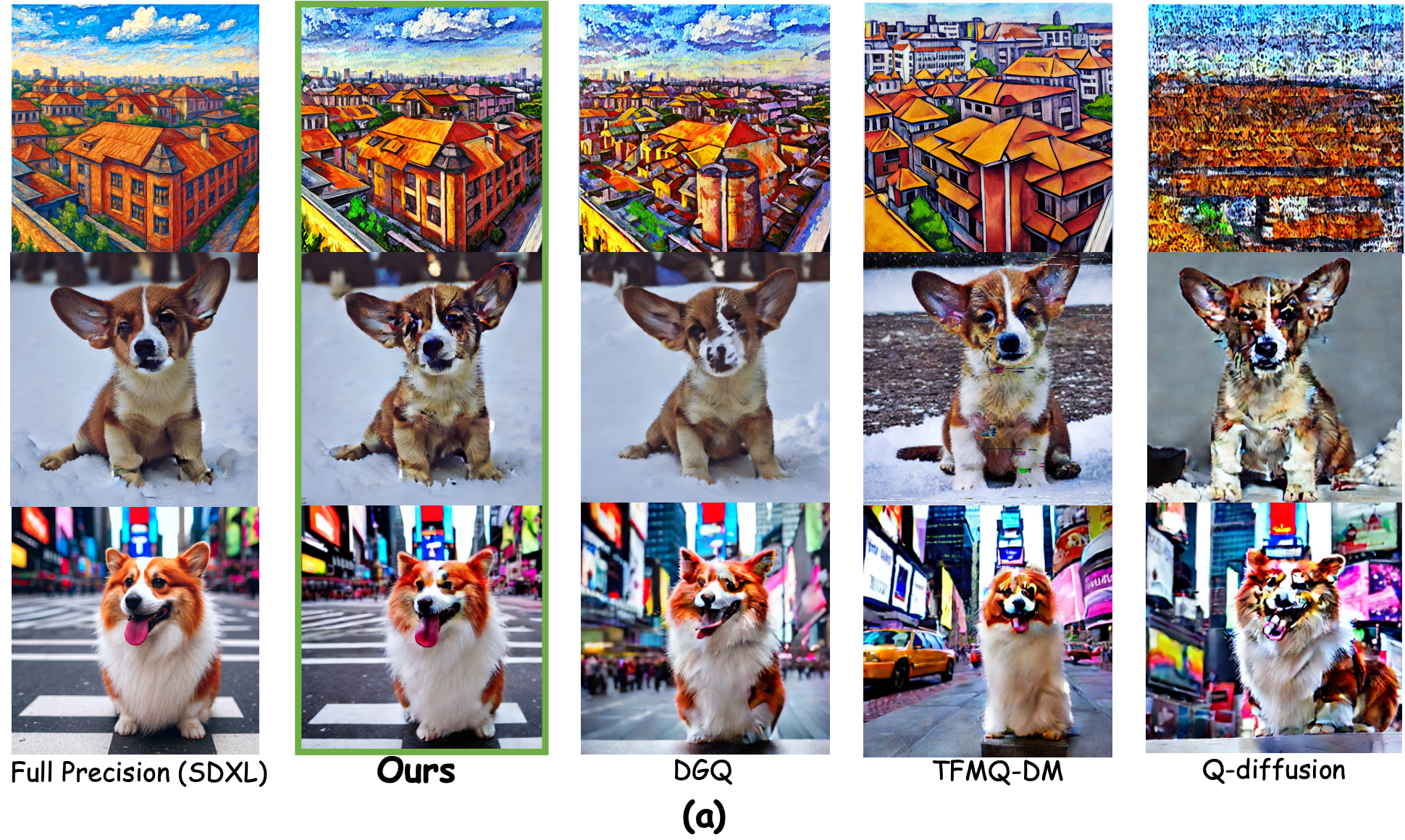}
\includegraphics[width=1\columnwidth]{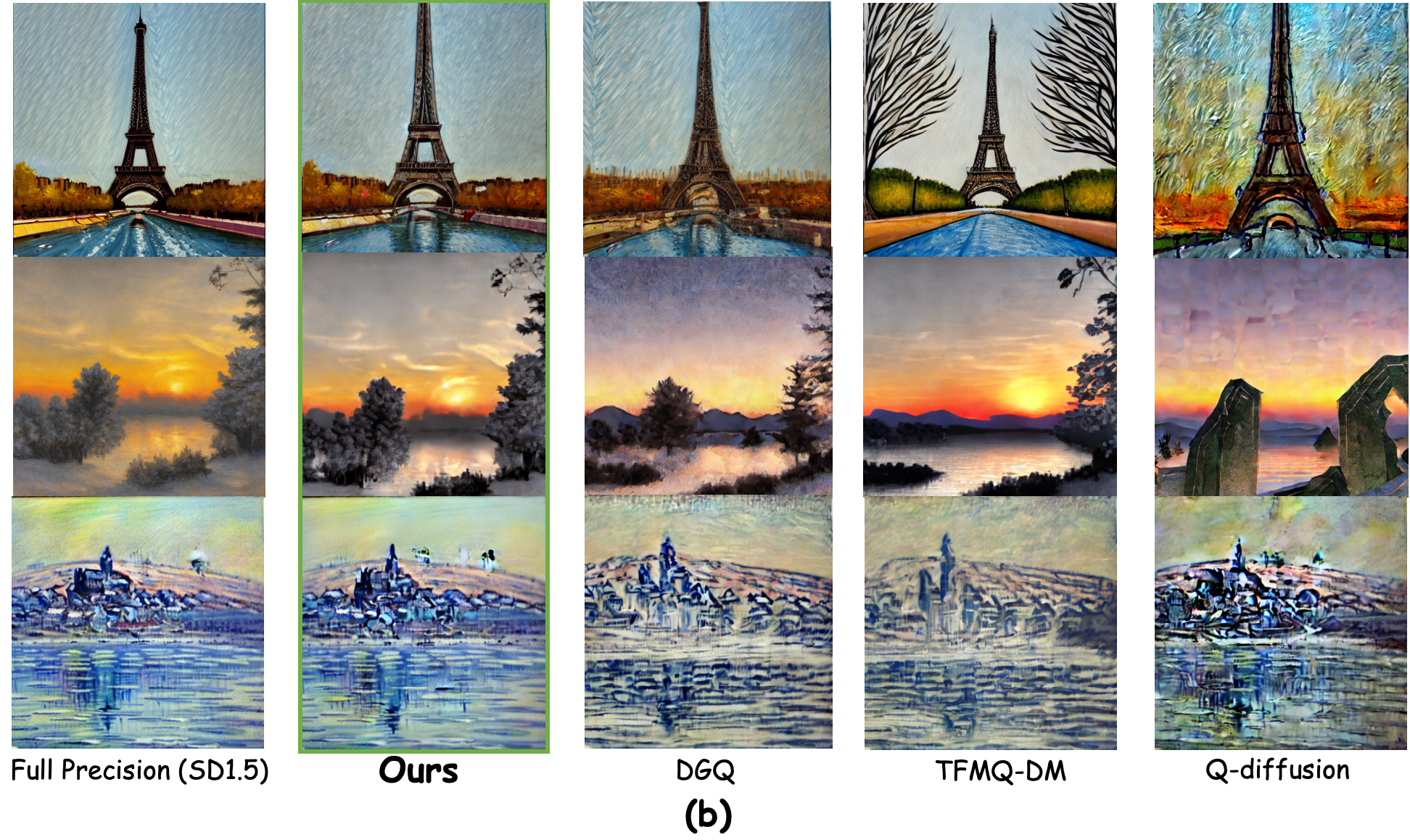}
\caption{\textbf{Qualitative comparison.}
(a) SDXL-Turbo with W8A8 (8-bit weights / 8-bit activations). (b) Stable Diffusion v1.5 with W8A8  (8-bit weights / 8-bit activations); Each column corresponds to a different
personalized checkpoint.}
\label{fig:qual_comp}
\label{fig:qual_comp}
\end{figure*}

\subsection{Quantitative Results}

\textbf{Automatic Benchmarks.}
We evaluate on MS-COCO and PartiPrompts using CLIP~\cite{radford2021learning} and FID~\cite{seitzer2020pytorch}. We also report computation via BOPs~\cite{yu2020search},
$\mathrm{BOPs}=\mathrm{FLOPs}\cdot b_w \cdot b_a$,
where $b_w$ and $b_a$ are weight/activation bit-widths. Table~\ref{tab:main} shows TAQ achieves the best quality--compression trade-off across backbones and datasets: at 8-bit it remains close to full precision, and at 4-bit activations it degrades far less than prior PTQ baselines. Similar gains hold for SDXL-Turbo, where TAQ avoids common failure cases.

\textbf{LLM as a Judge.}
We use an LLM judge to score checkpoint--request alignment on subject, style, temporal, and context fit (1--5 Likert each~\cite{gu2024survey}), and report \emph{Intent Score} as their average. To reduce positional bias~\cite{zheng2023judging}, we also score with flipped option orders and keep only consistent outcomes. Table~\ref{tab:checkpoint-selection} shows \emph{Check-in} achieves the highest \emph{Intent Score}.

\textbf{Human Evaluations.}
Using 500 prompts from \textsc{Repo-Prompts}, MS-COCO, and PartiPrompts, we compare Check-in vs each baseline (3 comparisons), totaling 1500 comparisons; each comparison is rated by one rater (randomly assigned), and we report \emph{Intent-Alignment Win Rate} (preference for \emph{Check-in}). As shown in Table~\ref{tab:checkpoint-selection}, \emph{Check-in} wins 89.1\% vs.\ Random, 85.7\% vs.\ Reranker, and 82.1\% vs.\ Stylus.

\begin{table}[t]
\caption{\textbf{Checkpoint selection.}
\textbf{Intent Score}: mean$\pm$std LLM-judge rating (1--5; temp=0) over subject/style/temporal/context.
\textbf{Human Preference} ($p$): fraction of pairwise votes where \emph{Check-in} is preferred over the method
($p=\Pr(\text{\emph{Check-in}} \succ \text{method})$; higher is better). We set $p=1.00$ for \emph{Check-in}.}
\label{tab:checkpoint-selection}
\vspace{-0.3em}
\centering
\small
\setlength{\tabcolsep}{5pt}
\renewcommand{\arraystretch}{1.05}
\begin{tabular}{lcc}
\toprule
\textbf{Selection Method} & \textbf{Intent Score}$\uparrow$ & \textbf{Human Pref.} $p\uparrow$ \\
\midrule
Random & 2.14 $\pm$ 0.82 & 0.891 \\
Reranker & 3.21 $\pm$ 0.76 & 0.857 \\
Stylus & 3.68 $\pm$ 0.69 & 0.821 \\
\rowcolor{taqgray}
\textbf{Check-in} & \textbf{4.42 $\pm$ 0.51} & \textbf{1.000} \\
\bottomrule
\end{tabular}
\vspace{-0.6em}
\end{table}

\subsection{Ablation Study}

\textbf{Effect of \emph{Check-in}.}
Table~\ref{tab:ablation_checkin} ablates two \emph{Check-in} modules: (1) \emph{intent-aware retrieval} and (2) \emph{personalized reasoning}. Clarification is constant across all variants and is invoked only when the intent parser detects ambiguity. Hybrid retrieval combines sparse matches over $S_i$ (type constraints) with dense similarity from $D_i$, yielding a stronger candidate set. Personalized reasoning provides a consistent boost by enforcing temporal preferences and version cues. Image-quality metrics change little across variants, since \emph{Check-in} primarily improves checkpoint selection rather than generation quality.

\begin{table}[t]
\caption{Effect of \emph{Check-in} components.
\textbf{Retr.}: intent-aware retrieval;
\textbf{Rsn.}: personalized reasoning. Clarification is enabled on-demand in all variants (held constant).}
\label{tab:ablation_checkin}
\vspace{-0.3em}
\centering
{\renewcommand{\arraystretch}{0.95}
\setlength{\tabcolsep}{4pt}
\small
\begin{tabular}{ccccc}
\toprule
\textbf{Retr.} & \textbf{Rsn.} & \textbf{Intent Score}$\uparrow$ & \textbf{FID}$\downarrow$ & \textbf{CLIP}$\uparrow$ \\
\midrule
$\times$ & $\times$ & 3.21 & 11.74 & 0.289 \\
$\times$ & \checkmark & 3.99 & 11.61 & 0.291 \\
\checkmark & $\times$ & 4.19 & 11.23 & 0.293 \\
\checkmark & \checkmark & \textbf{4.42} & \textbf{11.03} & \textbf{0.297} \\
\bottomrule
\end{tabular}}
\vspace{-0.6em}
\end{table}

\textbf{Effect of trigger-aware quantization (TAQ).} Table~\ref{tab:ablation_taq} shows one clear takeaway: separating trigger tokens matters. Without separation, quantization tends to damage fidelity, and the logarithmic quantizer is especially unstable. Once trigger tokens are handled separately, quantization becomes much more reliable, and at lower precision the logarithmic quantizer benefits the most and delivers the best results.

\begin{table}[t]
\caption{Effect of TAQ and trigger-token separation.}
\label{tab:ablation_taq}
\vspace{-0.3em}
\centering
{\renewcommand{\arraystretch}{0.95}
\setlength{\tabcolsep}{4pt}
\begin{tabular}{ccccc}
\toprule
\bfseries Bits(W/A) & \bfseries Quantizer & \bfseries Separate trigger tokens & \bfseries FID$\downarrow$ & \bfseries CLIP$\uparrow$ \\
\midrule
8/8 & Linear & $\times$ & 15.83 & 0.292 \\
8/8 & Linear & \checkmark & \textbf{11.04} & \textbf{0.298} \\
8/8 & Logarithmic & $\times$ & 17.62 & 0.287 \\
8/8 & Logarithmic & \checkmark & 13.67 & 0.294 \\
\midrule
8/4 & Linear & $\times$ & 54.12 & 0.245 \\
8/4 & Linear & \checkmark & 44.53 & 0.262 \\
8/4 & Logarithmic & $\times$ & 56.21 & 0.249 \\
8/4 & Logarithmic & \checkmark & \textbf{38.22} & \textbf{0.265} \\
\bottomrule
\end{tabular}
}
\vspace{-0.8em}
\end{table}

\textbf{Component synergy.} Table~\ref{tab:ablation_two_parts} confirms the two parts solve different problems. \emph{Check-in} primarily improves intent alignment by selecting more appropriate checkpoints, while TAQ primarily preserves image fidelity under quantization. This complementarity becomes most important at aggressive compression, where using both is the most robust option.

\begin{table}[t]
\caption{Effect of component synergy.}
\label{tab:ablation_two_parts}
\vspace{-0.3em}
\centering
{\renewcommand{\arraystretch}{0.95}
\setlength{\tabcolsep}{3pt}
\begin{tabular}{cccccc}
\toprule
\bfseries Bits(W/A) & \bfseries \emph{Check-in} & \bfseries TAQ & \bfseries Intent score$\uparrow$ & \bfseries FID$\downarrow$ & \bfseries CLIP$\uparrow$ \\
\midrule
\multirow{4}{*}{8/8}
& $\times$      & $\times$      & 3.22 & 15.22 & 0.286 \\
& \checkmark    & $\times$      & 4.39 & 14.81 & 0.294 \\
& $\times$      & \checkmark    & 3.21 & 11.74 & 0.289 \\
& \checkmark    & \checkmark    & \textbf{4.42} & \textbf{11.03} & \textbf{0.297} \\
\midrule
\multirow{4}{*}{8/4}
& $\times$      & $\times$      & 3.21 & 54.12 & 0.245 \\
& \checkmark    & $\times$      & 4.40 & 51.31 & 0.253 \\
& $\times$      & \checkmark    & 3.23 & 39.53 & 0.261 \\
& \checkmark    & \checkmark    & \textbf{4.41} & \textbf{38.22} & \textbf{0.265} \\
\bottomrule
\end{tabular}
}
\vspace{-0.8em}
\end{table}

\textbf{MLLM Backbones.}
To deploy on memory-constrained devices, we evaluated all backbones via API, avoiding local GPU requirements. We tested Qwen2.5-VL-72B, GPT-4o, and Gemini 2.5 Flash. Table~\ref{tab:mllm-ablation} shows similar intent scores across backbones, but latency differs: Gemini 2.5 Flash has the lowest end-to-end time, mainly due to faster reasoning and clarification, making it our preferred backbone under latency constraints.

\begin{table}[t]
\caption{\textbf{MLLM latency ablation.} Average number of user turns per request (including the initial prompt); $>1$ implies clarification.}
\label{tab:mllm-ablation}
\centering
\scriptsize
\setlength{\tabcolsep}{3pt}
\renewcommand{\arraystretch}{1.15}

\resizebox{\columnwidth}{!}{%
\begin{tabular}{lccccccc}
\hline
& \textbf{Performance} & \multicolumn{5}{c}{\textbf{Inference time (s)}} & \\
\cline{2-2}\cline{3-7}
\textbf{Backbone} &
\textbf{Intent Score} &
\textbf{Retrieve} & \textbf{Reason} & \textbf{Clarify} &
\textbf{Generation (W8A8)} & \textbf{End-to-End} &
\textbf{Multi-turn} \\
\hline
GPT-4o (API)           & 4.42 & 1.3 & 20.31 & 11.17 & 12.31 & 45.09 & 2.1 \\
Gemini 2.5 Flash (API) & 4.38 & 1.3 & 16.54 &  8.28 & 12.31 & 38.43 & 2.3 \\
Qwen2.5-VL-72B (API)   & 4.35 & 1.3 & 25.89 & 10.35 & 12.31 & 49.85 & 2.5 \\
\hline
\end{tabular}%
}
\end{table}

\section{Conclusion}
We introduced PersonalQ, a unified system for selecting and serving personalized diffusion models under tight memory budgets. Our \emph{Check-in} performs intent-aligned checkpoint selection by combining hybrid retrieval with personalized reasoning and brief clarification, then rewrites prompts with the selected checkpoint’s trigger token. Complementing this, \emph{TAQ} preserves trigger-conditioned cross-attention pathways while quantizing the remaining components, enabling memory-efficient inference with strong personalization fidelity. Together, these components support scalable deployment of large personalized checkpoint repositories without compromising intent alignment or generation quality.

\bibliographystyle{IEEEbib}
\bibliography{icme2026references}

\end{document}